\documentclass{article}

\usepackage[preprint]{neurips_2026}
\usepackage{amsmath}
\usepackage{amssymb}
\usepackage{mathtools}
\usepackage{amsthm}
\usepackage{algorithm}
\usepackage{algorithmic}
\usepackage[utf8]{inputenc} 
\usepackage[T1]{fontenc}    
\usepackage{hyperref}       
\usepackage{url}            
\usepackage{booktabs}       
\usepackage{amsfonts}       
\usepackage{nicefrac}       
\usepackage{microtype}      
\usepackage{xcolor}         
\definecolor{algorithmphase}{RGB}{35,80,150}

\usepackage[capitalize,noabbrev]{cleveref}
\usepackage[textsize=tiny]{todonotes}

\theoremstyle{plain}

\theoremstyle{definition}

\theoremstyle{remark}

\title{The Surprising Effectiveness of Approximate Value Iteration in Self-Play}

\author{Raphaël Boige\thanks{Corresponding author: \texttt{\{name\}.\{surname\}@inria.fr}} \qquad Amine Boumaza \qquad Bruno Scherrer\\Université de Lorraine, CNRS, Inria, LORIA, F-54000 Nancy, France}

\hypersetup{
    pdftitle={The Surprising Effectiveness of Approximate Value Iteration in Self-Play},
    pdfauthor={Raphaël Boige, Amine Boumaza, Bruno Scherrer}
}

\begin{document}
\maketitle
\begin{abstract}

Combining search with function approximation has driven major advances in game-playing programs, making self-play algorithms more competitive than ever. Still, the computational overhead of the most popular methods, based on Monte Carlo Tree Search (MCTS), can be substantial. In this work, we investigate whether simpler methods remain competitive in non-trivial, moderately sized games such as Connect Four, Hex(7x7) and synthetic games. We train a minimal self-play implementation of Approximate Value Iteration (AVI) and use ground-truth oracles for exact evaluation. Contrary to expectations, our results demonstrate the surprising effectiveness of AVI: it learns more accurate value functions than those learned by AlphaZero, while its one-step-lookahead greedy policies remain competitive with MCTS-based policies at substantially lower training and inference costs. Preliminary experiments on Othello and Go(9x9) show that AVI trains stably on larger games and learns effective value functions. These findings suggest that the success of MCTS-based methods may have eclipsed simpler approaches that have become increasingly practical with modern deep-learning tools.

\end{abstract}

\section{Introduction}

Two-player games extend classical sequential decision-making with an adversarial min-max structure: each player must account for the best response of its opponent. This makes planning and learning particularly challenging. Yet, the optimal value of a position can still be expressed recursively in terms of successor positions, much like the Bellman optimality equation in a Markov Decision Process (MDP). This conveniently allows the use of Reinforcement Learning (RL) or Approximate Dynamic Programming (ADP) methods and tools.

AlphaZero (AZ)~\citep{silver2018general} changed the way game-playing agents are built. Its success in Go, Chess, and Shogi made the combination of Monte Carlo Tree Search (MCTS) and deep neural networks a dominant paradigm. In AlphaZero, MCTS performs planning by using heuristics to direct simulations towards promising actions while preserving exploration~\citep{coulom2006efficient,kocsis2006bandit,rosin2011multi}. This mechanism is powerful but introduces both computational and engineering costs. It adds complex interactions between exploration/exploitation heuristics, policy learning and value learning. It also requires many neural-network calls at every move, during both training and inference. This raises a natural question: how much can be achieved by learning values directly, without tree search during training?

We revisit Approximate Value Iteration (AVI), a classical alternative that turns out to be particularly simple in alternating two-player games. Our implementation uses only $\epsilon$-greedy self-play, one-step negamax backups, and a neural value function. However, there are good reasons to expect this approach to fail. It combines function approximation, bootstrapping, and off-policy learning: the ingredients of the ``deadly triad''~\citep{sutton2018reinforcement} known for causing numerical instabilities or divergence. This combination has long made direct value learning appear risky and may partly explain why such methods received limited attention as MCTS-based approaches became dominant for games. Yet the expected failure mode does not materialize in our experiments: AVI trains stably across domains and random seeds while learning increasingly accurate value functions. Beyond stability, we study how accurate the learned values become, how well they support decision making, and where AVI's simple exploration and inference mechanisms reach their limits.

Evaluations of self-play algorithms usually rely on Elo ratings or head-to-head matches. These metrics measure relative playing strength against a particular set of opponents, but they provide no absolute reference and reveal little about the quality of the learned value functions. We therefore focus on games for which optimal values and actions are available: Connect Four, Hex(7x7), and synthetic F-Games~\citep{boige2025alphabeta}. The ground-truth oracles allow us to measure value error and policy regret exactly, in addition to evaluating play against a perfect opponent. To the best of our knowledge, this is the first systematic oracle-based comparison of neural AVI and MCTS-based self-play across multiple non-trivial games. The availability of exact oracles, however, restricts the size of the games we can study. We therefore complement this evaluation with experiments on larger games, Othello and Go(9x9), using MiniZero as an external AlphaZero baseline~\citep{wu2024minizero}.

Our main contributions are:
\begin{enumerate}
    \item We provide a controlled experimental comparison of AVI and AlphaZero on solved games, measuring value error, policy regret, play against perfect oracles, and computational cost.

    \item We show that AVI learns substantially more accurate value functions than our AlphaZero baseline. Its one-step-lookahead policies remain competitive at substantially lower cost, and its values provide stronger guidance for deeper search.

    \item We show that AVI trains stably on more complex domains and learns value functions that improve MiniZero's playing strength when used with the same MCTS inference budget.
\end{enumerate}

Our results suggest different insights for value learning and search-based action selection. Accurate values can be surprisingly well learned by direct negamax backups, while more advanced inference strategies (such as MCTS) remain valuable for turning those values into strong decisions.
\section{Related Work}

AlphaZero~\citep{silver2018general} established MCTS as a central policy-improvement mechanism for neural self-play. This role has also been formalized by interpreting MCTS as a form of regularized policy optimization~\citep{grill2020monte}. Many subsequent methods retain this structure while modifying other components. MuZero~\citep{schrittwieser2020mastering}, for example, learns the model used during planning, while Gumbel AlphaZero~\citep{danihelka2022policy} improves policy improvement under limited simulation budgets. Despite these differences, these methods all use search to produce an improved policy that subsequently guides learning.

Learning value functions in games predates AlphaZero. Early examples range from direct temporal-difference learning to agents that combine learned values with game-tree search~\citep{samuel1959some,tesauro1995temporal,baxter1999knightcap,schaeffer2001temporal}. In particular, \citet{veness2009bootstrapping} studied bootstrapped value learning from minimax search and obtained master-level Chess performance with linear function approximation. This line of work established that learned values and search can reinforce one another. It did not, however, isolate how far a minimal one-step value-iteration scheme can go with modern neural function approximation.

From an approximate dynamic programming perspective, value iteration and policy iteration can be viewed as endpoints of the broader approximate modified policy-iteration family~\citep{scherrer2012approximate}. Related analyses have extended approximate dynamic programming methods to two-player zero-sum Markov games~\citep{perolat2015approximate}. More recently, \citet{cohen2023minimax} revisited the learning of minimax values, while the Descent algorithm~\citep{cohen2026learning} used best-first minimax search to construct stronger training targets. Our work studies a more elementary approach: one-step negamax backups, $\epsilon$-greedy self-play, and no tree search during training. We complement this minimal design with oracle-based evaluation, which lets us measure learned values and decisions directly rather than relying only on relative playing strength.
\section{Problem Formulation and Algorithms}
\label{sec:problem_algorithms}

\subsection{Alternating Two-Player Zero-Sum Games}

We consider finite, deterministic, perfect-information games in which two players act in turn and receive opposite rewards. Let $\mathcal{S}$ be the state space, $\mathcal{A}(s)$ the legal actions in state $s$, and $f(s,a)$ the state reached after action $a$. We use the negamax convention: values and rewards are always expressed from the perspective of the player to move. This removes the need for separate maximization and minimization equations. The optimal value satisfies
\begin{equation}
    \label{eq:bellman-negamax}
    V^*(s) = \max_{a \in \mathcal{A}(s)}
    \left[ R(s,a) - \gamma V^*(f(s,a)) \right],
\end{equation}
where $\gamma \in [0,1]$ is the discount factor and terminal transitions carry the game outcome. For any value estimate $V$, the corresponding one-step-lookahead greedy policy is
\begin{equation}
    \label{eq:greedy-policy}
    \mathcal{G}(V)(s) \in \arg\max_{a \in \mathcal{A}(s)}
    \left[ R(s,a) - \gamma V(f(s,a)) \right].
\end{equation}
A more general Markov-game formulation is provided in Appendix~\ref{appendix:algorithm_details}.

\subsection{Value Iteration in Games}

The negamax equation suggests a direct algorithm for solving the game. Starting from an initial value function $V_0$, Value Iteration (VI) repeatedly applies the optimality operator to every state:
\begin{equation}
    \label{eq:value-iteration}
    V_{k+1}(s)
    = \max_{a\in\mathcal{A}(s)}
    \left[R(s,a)-\gamma V_k(f(s,a))\right].
\end{equation}
In an alternating game of horizon $h$, exact VI recovers the optimal values after at most $h$ iterations. The difficulty is computational: each iteration requires updating every state and enumerating all of its legal actions, which is impossible in the large state spaces that interest us.

\subsection{Approximate Value Iteration}

Approximate Value Iteration (AVI) replaces the table of values with a function approximator (in our case, a neural value function $V_\theta$) and the exhaustive state-space update with regression on sampled states. Given parameters $\bar\theta$ held fixed during data collection, we compute one-step negamax targets
\begin{align}
    y_{\bar\theta}(s)
    &= \max_{a \in \mathcal{A}(s)}
    \left[R(s,a)-\gamma V_{\bar\theta}(f(s,a))\right],
    \label{eq:avi-target}
\end{align}
and fit the function approximator using a mean-squared error loss (MSE):
\begin{align}
\mathcal{L}_{\mathrm{AVI}}(\theta)
    &= \mathbb{E}_{(s,y)\sim\mathcal{D}}
    \left[\left(V_\theta(s)-y\right)^2\right].
    \label{eq:avi-loss}
\end{align}
The choice of the sampling distribution $\mathcal{D}$ is central to AVI. Exact VI updates the whole state space, whereas AVI only improves the value function on the distribution represented in its training data. A narrow distribution may miss strategically important states; a broad but uninformed distribution may spend most of its budget on irrelevant ones. We generate this distribution online through $\epsilon$-greedy self-play with respect to Equation~\ref{eq:greedy-policy}. This concentrates data around the current strategies while retaining some exploration. We store the resulting state-target pairs in a circular replay buffer, which mixes data collected at different stages of learning.

Training alternates between data collection, during which the network is fixed, and optimization on the replay buffer. Algorithm~\ref{alg:avi} gives the complete procedure.

\begin{algorithm}[H]
    \caption{Approximate Value Iteration for alternating zero-sum games}
    \label{alg:avi}
    \begin{algorithmic}[1]
        \REQUIRE iterations $I$, environments $N_{\mathrm{envs}}$, rollout steps $K$, gradient steps $N_{\mathrm{grads}}$, batch size $B$, exploration rate $\epsilon$, discount factor $\gamma$
        \STATE Initialize value network $V_\theta$, circular replay buffer $\mathcal{D}$, and $N_{\mathrm{envs}}$ parallel environments
        \FOR{$i=1,\ldots,I$}
            \STATE Set $\bar\theta \leftarrow \theta$
            \STATE \textcolor{algorithmphase}{\texttt{// 1. Data Collection Phase}}
            \FOR{$k=1,\ldots,K$}
                \FOR{each parallel environment with state $s$}
                    \STATE Compute $q(s,a)=R(s,a)-\gamma V_{\bar\theta}(f(s,a))$ for every $a\in\mathcal{A}(s)$
                    \STATE Store the state-target pair $(s,\max_a q(s,a))$ in $\mathcal{D}$
                    \STATE Select $a$ $\epsilon$-greedily from $q$ and step forward the environment (resetting it if terminal)
                \ENDFOR
            \ENDFOR
            \STATE \textcolor{algorithmphase}{\texttt{// 2. Training Phase}}
            \FOR{$g=1,\ldots,N_{\mathrm{grads}}$}
                \STATE Sample a mini-batch of $B$ state-target pairs from $\mathcal{D}$
                \STATE Update $\theta$ by minimizing the mean-squared error in Equation~\ref{eq:avi-loss}
            \ENDFOR
        \ENDFOR
    \end{algorithmic}
\end{algorithm}

\subsection{AlphaZero Baseline}

Our AlphaZero baseline follows the standard self-play procedure~\citep{silver2018general}. Its network predicts both a value $V_\theta(s)$ and a policy $\pi_\theta(a\mid s)$. During data collection, MCTS uses these predictions to run $S$ simulations and returns an improved policy $\pi_{\mathrm{MCTS}}$ from its visit counts. This policy is recorded as a target at every visited state. Once the episode terminates, its final outcome $z$ provides the value target for all states visited during that episode. The network is then trained to predict both targets. Thus, AVI and AlphaZero differ not only in action selection, but also in their value targets: AVI learns from one-step negamax backups, whereas AlphaZero learns from complete self-play outcomes. Detailed MCTS equations and AlphaZero pseudocode are given in Appendix~\ref{appendix:algorithm_details}.

\section{Experimental Setup}

We use two complementary evaluation regimes. Our primary study focuses on Connect Four, Hex(7x7), and synthetic F-Games, where exact values and optimal actions are available. These oracles let us evaluate the learned values and decisions independently of a particular opponent. We then study Othello and Go(9x9), which are substantially larger but cannot be evaluated exactly. In these games, we measure playing strength against a fixed reference, MiniZero~\citep{wu2024minizero}, and assess the utility of AVI's value function when it is used within MiniZero's search.

\subsection{Games and Evaluation Regimes}
\label{sec:games_evaluation}

\paragraph{Connect Four.} Connect Four is played on a 6x7 grid and has approximately $10^{12}$ possible positions. Its branching factor is at most 7, but the game remains strategically deep. We use a perfect alpha-beta solver that evaluates every position in less than $1$sec and most positions in less than $0.1$sec. It returns both the theoretical outcome and the number of moves to termination, allowing it to select the quickest win or delay a forced loss; we therefore call it a \emph{strong oracle}. The rules and solver are detailed in Appendix~\ref{appendix:connect_four}.

\paragraph{Hex(7x7).} Hex is a connection game with a much wider branching factor: up to 49 legal actions on an empty 7x7 board. Its state space contains approximately $10^{22}$ positions and is too large for a generic minimax solver. We use a perfect domain-specific solver based on virtual connections and topological properties. It returns only the theoretical outcome, so we call it a \emph{weak oracle}. It remains exact for wins and losses but does not rank actions with the same outcome, and therefore plays arbitrarily when every move loses. See Appendix~\ref{appendix:hex} for details.

\paragraph{Synthetic F-Games.} F-Games~\citep{boige2025alphabeta} are procedurally generated game trees whose optimal values are known by construction. Their depth, branching factor, and difficulty can be controlled independently. They have no spatial state representation: a state is identified only by the sequence of actions leading to it. Generalization is therefore difficult, while exploration can be studied without spatial inductive biases. We consider three tree structures and generate 10 games for each. Their construction is described in Appendix~\ref{appendix:f_games_generation}.

\paragraph{Othello.} Othello is played on an $8\times8$ board. Starting from four occupied squares, a game can contain up to 60 placements, compared with 42 moves in Connect Four and 49 in Hex(7x7). The legal-action set varies with the position and has a typical size of 10.

\paragraph{Go(9x9).} Go(9x9) is played on 81 intersections. At the start of the game, the player can choose among 81 placements in addition to passing. Captures also allow games to last longer than the number of board intersections. Go(9x9) therefore combines a larger branching factor with longer-term strategic dependencies and is more broadly used by the community as a testbed for self-play algorithms.

\paragraph{Evaluation without oracles.} Neither game admits the kind of practical exact oracle available for Connect Four and Hex(7x7). We therefore evaluate AVI through games against the published MiniZero agents. This second regime does not provide an absolute measure of optimality, but tests whether AVI trains stably and learns values that remain useful at a larger scale.

\subsection{Oracle-Based Metrics}
\label{sec:metrics}

The exact oracles allow us to evaluate three different aspects of an agent: its value estimates, its immediate decisions, and its behavior over a complete game.

\paragraph{Value function error.} We compute the mean absolute error (MAE) between the learned value function and the ground truth on a fixed set of states $\mathcal{D}_\text{eval}$:
\begin{equation}
    \label{eq:mae}
    \operatorname{MAE}(V_\theta)
    = \mathbb{E}_{s\sim\mathcal{D}_\text{eval}}
      \left|V^*(s)-V_\theta(s)\right|.
\end{equation}
This metric directly quantifies the accuracy of the value function.

\paragraph{Policy regret.} Value accuracy alone is not a guarantee of the strength of the implicit greedy policy (e.g. a perfect value function with a constant bias would look inaccurate but would still yield an optimal greedy policy). To evaluate decisions, we measure the loss incurred by choosing an action from a policy $\pi$ instead of an optimal action:
\begin{equation}
    \label{eq:regret}
    \operatorname{Regret}(\pi_\theta)
    = \mathbb{E}_{s\sim\mathcal{D}_\text{eval}}
      \left[V^*(s)-\left(R(s,a)-\gamma V^*(s')\right)\right],
\end{equation}
where $a=\pi_\theta(s)$ and $s'=f(s,a)$. A policy has zero regret exactly when it selects an optimal action on every evaluated state. Since AVI does not learn a policy explicitly, we evaluate $\mathcal{G}(V_\theta)$ unless specified otherwise.

\paragraph{Play against an oracle.} Finally, we play complete games against the perfect solver from a fixed set of openings $\mathcal{D}_\text{openings}$. The \emph{error rate} is the fraction of games in which the agent obtains a worse outcome than the theoretical value of the opening: for example, turning a guaranteed win into a draw or loss. For Connect Four, we also report the \emph{blunder rate}, which counts direct transitions from a guaranteed win to a guaranteed loss. This test is stricter than one-step regret because a perfect opponent exploits every consequential mistake.

\subsection{Match-Based Evaluation}
\label{sec:match_evaluation}

When no oracle is available, we report the average game score, assigning $+1$ to a win, $0$ to a draw, and $-1$ to a loss. Every opening is played twice with the players' roles exchanged. This controls for first-player advantage and gives a score in $[-1,1]$.

We use the same protocol for direct comparisons between AVI and AlphaZero on the solved games, and between AVI and MiniZero on Othello and Go(9x9). Greedy AVI selects actions with $\mathcal{G}(V_\theta)$. AlphaZero and MiniZero select actions with MCTS at the stated simulation budget.

We call \emph{cross-inference} the configuration in which MCTS keeps the policy prior learned by AlphaZero or MiniZero but evaluates leaf nodes with AVI's value function. Thus, for example, we compare native inference using $(\pi_\text{MiniZero},V_\text{MiniZero})$ with cross-inference using $(\pi_\text{MiniZero},V_\text{AVI})$. This isolates the contribution of the value function while keeping the search procedure and number of simulations fixed. Cross-inference is a diagnostic experiment, not a practical hybrid agent: because the policy and value come from different networks, it requires two backbone evaluations and is therefore more expensive in wall-clock time.

\subsection{Evaluation States and Openings}

Table~\ref{tab:evaluation_datasets} summarizes the fixed state and opening sets. The oracle-based state sets are used for MAE and regret. The opening sets seed games against the oracle and direct head-to-head matches. Each match is played from both sides of every opening.

\begin{table}[t]
    \centering
    \small
    \caption{Evaluation states, openings, and reference used in each game.}
    \label{tab:evaluation_datasets}
    \begin{tabular}{llll}
        \toprule
        Game & Evaluation states & Openings & Reference \\
        \midrule
        Connect Four & $2\times1500$ & 406 (after 4 plies) & Perfect solver (strong)\\
        Hex(7x7) & 445 & 128 (after 2 plies) & Perfect solver (weak)\\
        F-Games & 4096 per game & -- & Known by construction \\
        Othello & -- & 54 (after 3 plies) & MiniZero \\
        Go(9x9) & -- & 81 (after 1 ply) & MiniZero \\
        \bottomrule
    \end{tabular}
\end{table}

For Connect Four, the two evaluation sets contain 1500 states sampled respectively through uniform and $\epsilon$-optimal play. They are stratified by game phase and difficulty. The Hex(7x7) evaluation states are sampled uniformly, while its opening set contains 128 positions after two plies, balanced evenly between the two solver outcomes. For each F-Game, we sample 4096 states uniformly. The Othello opening set contains all 54 positions attainable after three plies, and the Go(9x9) set contains the 81 positions attainable after the first move. Full generation details are provided in Appendices~\ref{appendix:connect_four}, \ref{appendix:hex}, \ref{appendix:othello_go}, and~\ref{appendix:f_games_generation}.

\subsection{Training and Comparison Protocol}
\label{sec:training_protocol}

\paragraph{Shared implementation.} We implement AVI and AlphaZero in JAX~\citep{bradbury2018jax} and reuse the same environments, replay buffer, optimization code, and neural-network backbone. AlphaZero only adds a policy head. Our implementation is adapted from PGX~\citep{koyamada2023pgx} and uses \textit{mctx}~\citep{deepmind2020jax} for MCTS. The environments and training loops execute almost entirely on the GPU. Our code is publicly available.\footnote{\url{https://github.com/Egiob/avi}}

\paragraph{Training and compute budgets.} On Connect Four and Hex(7x7), both methods collect the same number of training rows and ingest the same number of examples during optimization. They use the same batch size and number of gradient updates. Constructing these rows has a different cost: AVI evaluates every one-step successor, whereas AlphaZero performs $S$ tree simulations. We therefore report a forward-equivalent neural-network compute proxy, defined as $N_\text{forward}+3N_\text{backward}$. The factor of three approximates the greater cost of a backward pass. At inference, where no backward pass is required, this reduces to the number of forward calls. Figure~\ref{fig:az_vs_vi} reports this quantity relative to AVI for training and inference. This proxy captures the main computational difference without relying on wall-clock measurements from heterogeneous machines; it should not be interpreted as elapsed time.

\paragraph{Hyperparameters.} We run $N_\text{envs}=128$ parallel environments and collect $K=128$ steps per iteration. We tune the main exploration parameter of each method. For AVI, we search over the $\epsilon$-greedy exploration rate and select $\epsilon=0.3$. For AlphaZero, we tune $c_\text{puct}$ for simulation budgets $S\in\{32,64,128,256,512\}$ and use $c_\text{puct}=3$, which performs best on average. Architectures, optimization parameters, replay-buffer sizes, and the complete search spaces are reported in Appendix~\ref{appendix:experimental_details}.

\paragraph{External AlphaZero baselines.} We use two external implementations for different purposes. On Connect Four, AlphaZero.jl provides an independent calibration for the strength of our in-house AlphaZero agent. Because its architecture, training procedure, and compute budget differ from ours, we report this comparison separately in Appendix~\ref{appendix:alphazero_jl} rather than include it in the controlled main comparison. On Othello and Go(9x9), MiniZero provides the reference networks and opponents. We match its network architecture when training AVI, use $\epsilon=0.15$ to account for the larger branching factors, and leave the rest of the AVI procedure unchanged. MiniZero uses 200 MCTS simulations per move in our evaluations, matching its training budget.

\subsection{Statistical Reporting}

Every AlphaZero configuration on Connect Four and Hex(7x7) is trained with five random seeds. We use 20 seeds for AVI on these games to examine its stability more precisely. The Othello and Go(9x9) experiments use five AVI seeds, while F-Game results are aggregated over 10 independently generated games per tree structure. Unless stated otherwise, we report means and 95\% confidence intervals across independent runs using Student's $t$ distribution. Direct AVI vs AlphaZero matches use the five paired final checkpoints available for both methods.

\section{Results}

We organize the results around three questions. Does AVI learn accurate values without becoming unstable? Do these values support strong decisions, with and without deeper search? Finally, does the same learning procedure remain effective on larger games? Connect Four and Hex(7x7) provide the most controlled evidence because every evaluation state can be solved exactly. F-Games provide a complementary synthetic check, while Othello and Go(9x9) test the method at a larger scale.

\subsection{Accurate Values and Competitive Policies under Exact Evaluation}
\label{sec:vi_vs_az}

\begin{figure}[H]
    \centering
    \includegraphics[width=1.\linewidth]{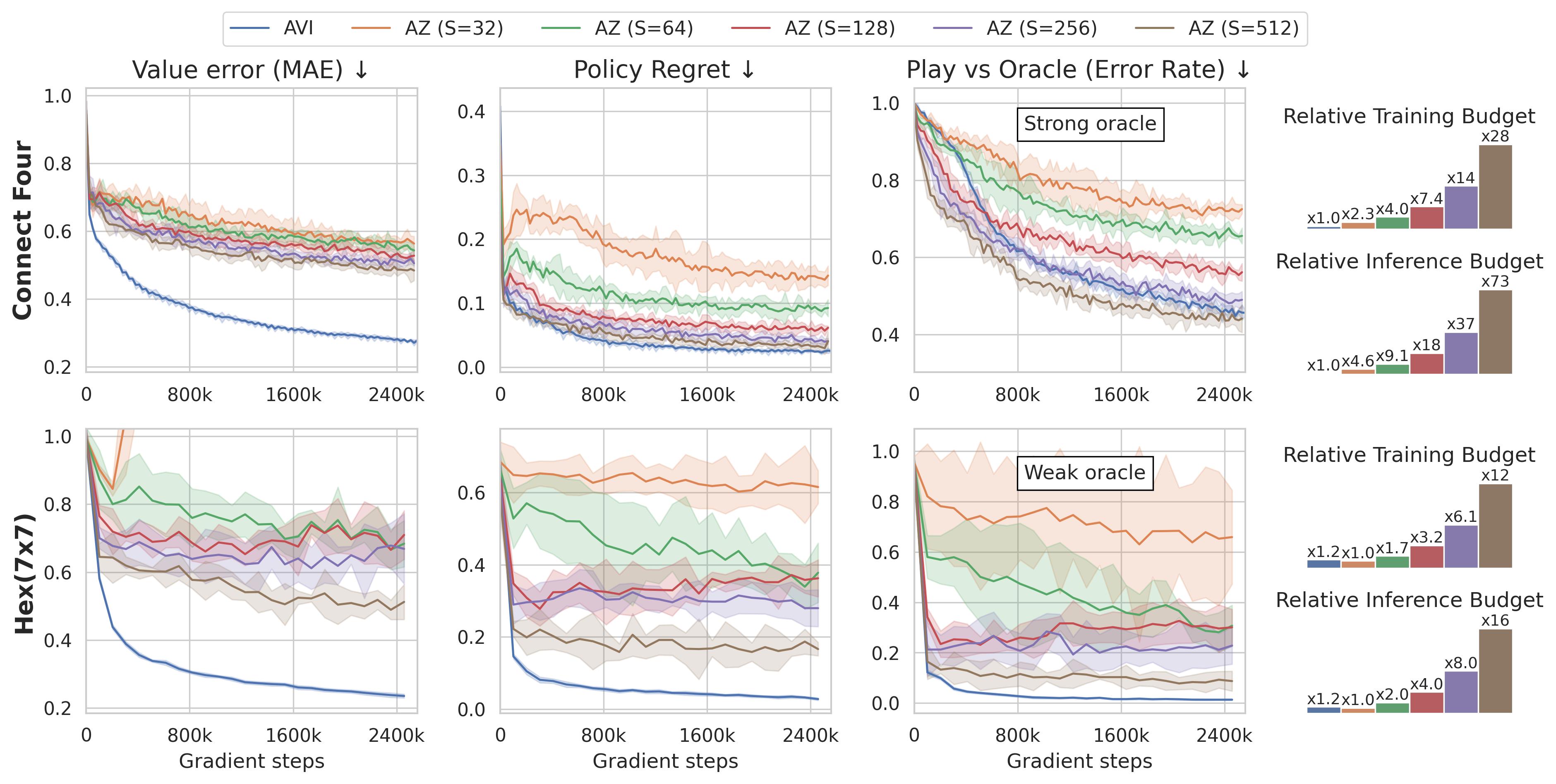}
    \caption{AVI and AlphaZero on Connect Four (top) and Hex(7x7) (bottom). From left to right: value error, policy regret, error rate against the oracle, and relative forward-equivalent compute during training and inference. Connect Four uses the strong oracle and Hex(7x7) the weak oracle defined in Section~\ref{sec:games_evaluation}. We report means over 20 seeds for AVI and 5 seeds for AZ. Shaded regions show 95\% confidence intervals computed using Student's $t$ distribution. ↓ indicates that lower is better.}
    \label{fig:az_vs_vi}
\end{figure}

We first present the results on Connect Four and Hex(7x7). Figure~\ref{fig:az_vs_vi} follows AVI and AlphaZero across the tested search budgets $S\in \{32, 64, 128, 256, 512\}$. Both methods are trained for $2.4$M gradient updates under the common data and optimization protocol described in Section~\ref{sec:training_protocol}.

\paragraph{Value accuracy and stability.} AVI reaches a lower value error in both games, and the gap grows over training. Also, as expected, increasing AlphaZero's MCTS budget improves its value estimates. Yet even at its largest search budget, its value error remains higher than AVI's. Perhaps most notably, AVI learns smoothly across all 20 runs: despite combining function approximation, bootstrapping, and off-policy learning, we observe no divergence in either game.

\paragraph{Decision quality.} Despite the simplicity of the greedy policy, the advantage in value seems to carry over to one-step decisions. On Connect Four, greedy AVI reaches a regret and error rate close to the strongest AlphaZero configurations. On Hex(7x7), it obtains both lower regret and a lower error rate than every AlphaZero budget. As an external calibration, Figure~\ref{fig:alphazero_jl} compares the Connect Four results with AlphaZero.jl; the independently developed agent performs on the same scale as our strongest in-house AlphaZero configuration.

The oracle metrics are complemented by head-to-head matches between the final agents. As shown in Table~\ref{tab:avi_az_hth}, greedy AVI performs similarly or better compared to our strongest AlphaZero variant ($S=512$).

\begin{table}[H]
    \centering
    \small
    \caption{Head-to-head score of greedy AVI against AlphaZero with 512 MCTS simulations. Positive scores favor AVI. We report 95\% confidence intervals across the five paired final checkpoints.}
    \label{tab:avi_az_hth}
    \begin{tabular}{lc}
        \toprule
        Game & AVI score \\
        \midrule
        Connect Four & $-0.09 \pm 0.04$ \\
        Hex(7x7) & $0.38 \pm 0.09$ \\
        \bottomrule
    \end{tabular}
\end{table}

We find this performance to be especially notable in view of the compute comparison. AVI constructs its targets without tree search and uses only one-step lookahead at inference. When the strongest AZ agent requires 512 neural network evaluations per decision, AVI only requires a maximum of 7 neural network evaluations for Connect Four and 49 for Hex(7x7). This budget mismatch is illustrated in the last column of Figure~\ref{fig:az_vs_vi}.

\paragraph{F-Games.} We use F-Games as a complementary check across deep and narrow, balanced, and shallow and wide trees. AVI obtains lower value error across these structures, although the difference in policy quality is smaller on deep and narrow trees, where MCTS can compensate for inaccurate leaf values. We leave the detailed results to Appendix~\ref{appendix:f_games_generation} and Figure~\ref{fig:fgame_az_vs_vi}.

\subsection{AVI Values Provide Stronger Search Guidance}

\begin{figure}[h]
    \centering
    \includegraphics[width=\linewidth]{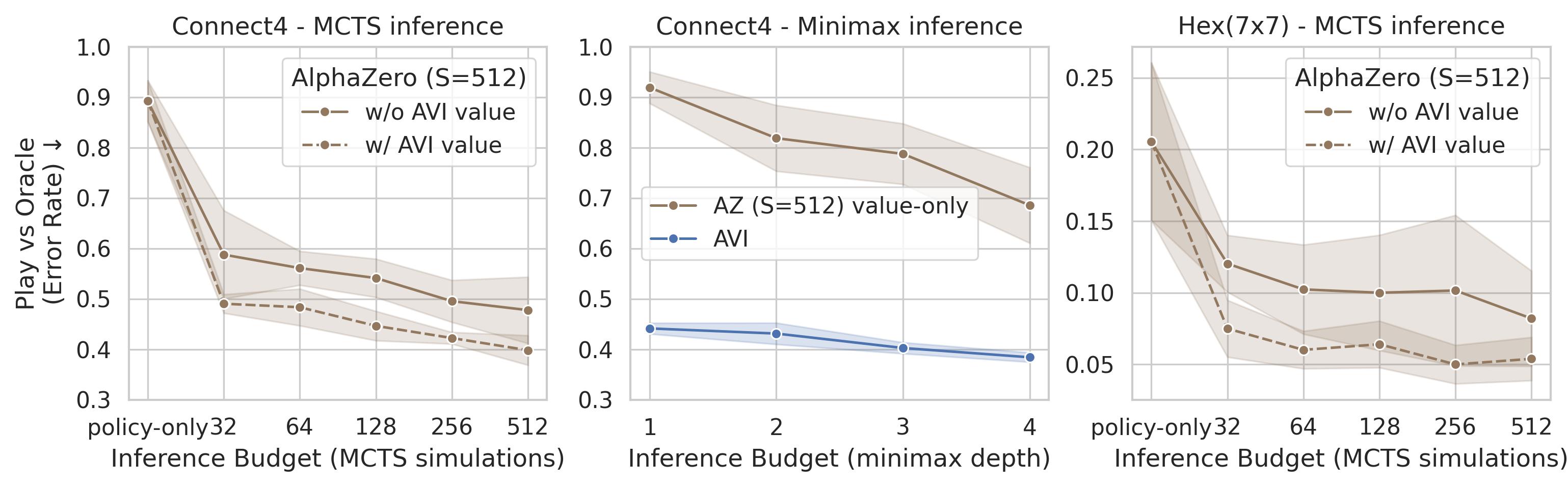}
    \caption{Inference with the values learned by AVI and AlphaZero. Left: error rate on Connect Four in cross-inference, i.e. when MCTS uses the AlphaZero policy with either its native value or AVI's value. Middle: value-only minimax inference on Connect Four. Right: the same MCTS cross-inference comparison on Hex(7x7). Curves average five seeds and shaded regions show 95\% confidence intervals. ↓ indicates that lower is better.}
    \label{fig:cross_inference}
\end{figure}

The oracle metrics establish that AVI learns more accurate values and competitive greedy policies. We next ask whether this value advantage persists when the value functions are embedded in more advanced search procedures. Figure~\ref{fig:cross_inference} reports two such tests: cross-inference, defined in Section~\ref{sec:match_evaluation}, and value-only minimax lookahead.

\paragraph{Cross-inference.} With the AlphaZero policy held fixed, replacing its value by AVI's value reduces the error rate throughout the range of MCTS budgets in both games. This shows that AVI's values integrate well into a more advanced inference scheme and provide better guidance for the same policy-guided search.

\paragraph{Value-only minimax.} Removing the policy network gives the same picture: both value functions benefit from deeper lookahead, but AVI remains markedly stronger. This suggests that AlphaZero relies more heavily on policy-guided search to compensate for its less accurate value estimates. Together, these results show that AVI's greater value accuracy translates into stronger decisions under different inference procedures.

\subsection{Scaling to Othello and Go(9x9)}

\begin{figure}[H]
    \centering
    \includegraphics[width=0.8\linewidth]{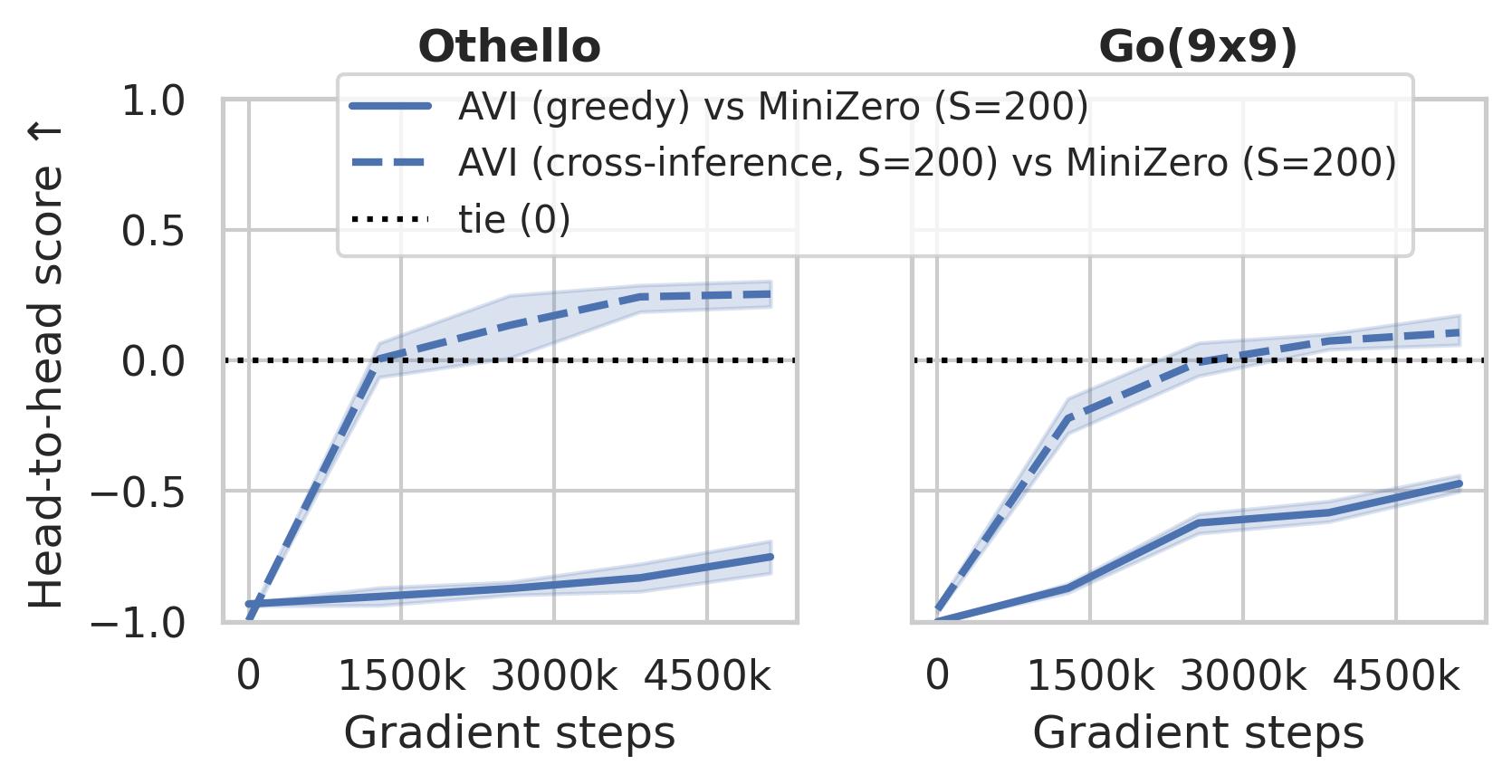}
    \caption{Head-to-head score against MiniZero with 200 MCTS simulations during AVI training on Othello (left) and Go(9x9) (right). Solid lines use greedy AVI. Dashed lines use MiniZero's policy and AVI's value under the same 200-simulation MCTS budget. Each opening is played from both sides. Curves average five AVI seeds and shaded regions show 95\% confidence intervals computed with Student's t distribution. ↑ indicates that higher is better.}
    \label{fig:go_othello_hth}
\end{figure}

Exact oracles make exact evaluation possible, but they also restrict the size of the games we can study, somewhat limiting the breadth of our conclusions. We therefore test the unchanged AVI learning rule on Othello and Go(9x9). Figure~\ref{fig:go_othello_hth} reports head-to-head scores against MiniZero, both for greedy AVI and for cross-inference under the same MCTS budget.

\paragraph{Greedy inference.} The same stable learning behavior persists on Othello and Go(9x9), suggesting that it is not confined to games for which exact evaluation is available. Indeed, AVI's greedy policy improves steadily, although it remains weaker than MiniZero's search-based agent in direct head-to-head matches. This comparison demonstrates a limitation of greedy inference in complex domains: it evaluates every legal action once and cannot concentrate additional computation on promising lines.

\paragraph{Cross-inference.} The learned values are nevertheless useful: when used within MiniZero's MCTS, they improve its playing strength in both games. We find this result especially striking: despite its simple training procedure, AVI learns value functions that improve an independently trained MCTS agent under the same search budget.

\section{Discussion and Limitations}

In this work, we investigated how far a minimal Approximate Value Iteration scheme could go when trained with modern deep-learning tools. Using oracle-based evaluation on Connect Four, Hex(7x7), and F-Games, we found that AVI learns substantially more accurate value functions than our AlphaZero baseline, and does so stably across random seeds. These more accurate values also translate into competitive greedy policies, despite far smaller training and inference budgets. On the larger games of Othello and Go(9x9), where exact values are unavailable, AVI again trains stably, and its learned values improve MiniZero's playing strength when inserted into the same MCTS inference procedure.

This stability is itself a central finding, and one plausible explanation lies in the alternating structure of the games themselves. In MDPs, maximizing over noisy value estimates can introduce a systematic overestimation bias~\citep{hasselt2010double}. In a negamax backup, however, successor values are instead expressed from the opponent's perspective and enter with the opposite sign, so an overestimate for one player becomes an underestimate one ply earlier: approximation errors may therefore alternate in sign rather than accumulate in the same direction. This bias-cancellation argument remains a hypothesis, but it suggests that error propagation in alternating games may behave better than in the usual MDP setting. Our experiments only show that the ingredients of the ``deadly triad'' do not necessarily lead to divergence under our training procedure and in the domains considered here.

The cross-inference results also help clarify what it means for AVI to be a simpler alternative to AlphaZero. Its main advantage lies at training time, where it learns accurate values without relying on deep search to generate either actions or targets. This advantage does not, however, require greedy one-step lookahead to remain the policy used at deployment: as the cross-inference experiments show, on games with many legal actions the same learned values can benefit from a more selective search procedure such as MCTS. Training and inference can therefore be designed separately, and simple, direct value learning may still support richer action selection when it is needed.

These conclusions come with clear limits. Our strongest claims rest on exact evaluation and therefore concern Connect Four, Hex(7x7), and F-Games; the Othello and Go(9x9) experiments, by contrast, are benchmarked only against a fixed MiniZero baseline and cannot reveal how far either agent is from optimal play. We also have not tested games at the scale of Go(19x19) or Chess. On the computational side, our forward-equivalent proxy captures the main neural-network cost but not every source of wall-clock overhead, such as building and populating the MCTS trees.

These limits suggest several directions for future work. Better exploration and state sampling may help AVI discover sparse strategies that $\epsilon$-greedy self-play could miss. Doubly-Asynchronous Value Iteration (DAVI)~\citep{tian2022doubly} provides a promising starting point for games with many legal actions: it replaces exhaustive maximization with maximization over a sampled subset of actions. Adapting this principle to neural self-play could address the weakness observed on Othello and Go(9x9), where greedy AVI spreads its budget evenly across every legal action; a learned policy prior could then guide action sampling toward promising moves. The cross-inference results point to a further, practical direction: a hybrid agent that learns AVI-style values and such a policy prior within a shared representation. Finally, our results do not show that MCTS is unnecessary; rather, they suggest that search may be more important for allocating inference-time computation than for learning useful values. Direct value learning might therefore deserve renewed attention, both on its own and as a component of search-based agents.

\section*{Acknowledgments}
This work was supported by a PhD grant from Université de Lorraine. This work was granted access to the HPC resources of IDRIS under the allocation 2026-AD011017585 made by GENCI. Some experiments presented in this paper were carried out using the Grid'5000 testbed, supported by a scientific interest group hosted by Inria and including CNRS, RENATER and several Universities as well as other organizations (see https://www.grid5000.fr).
\bibliography{example_paper}

\appendix

\section{Additional Formulation and Algorithm Details}
\label{appendix:algorithm_details}

\subsection{General Markov-Game Formulation}

A two-player Markov game is defined by a tuple $(\mathcal{S},\mathcal{A}_1,\mathcal{A}_2,P,R_1,R_2,\gamma)$, where $\mathcal{S}$ is the state space, $\mathcal{A}_i$ is the action space of player $i$, $P$ is the transition function, $R_i$ is the reward of player $i$, and $\gamma\in[0,1]$ is the discount factor. In a zero-sum game, $R_1=-R_2$. In an alternating game, the state space can be partitioned into $\mathcal{S}_1$ and $\mathcal{S}_2$, according to the player who acts. Policies $\pi_i:\mathcal{S}_i\rightarrow\Delta(\mathcal{A}_i)$ induce the value
\begin{equation}
    V^{\pi_1,\pi_2}(s)
    = \mathbb{E}\left[
        \sum_{t=0}^{\infty}\gamma^t R(s_t,a_t)
        \mid s_0=s,\pi_1,\pi_2
    \right].
\end{equation}
Solving the game amounts to finding policies that attain the minimax value
\begin{equation}
    V^{\pi_1^*,\pi_2^*}
    = \max_{\pi_1}\min_{\pi_2}V^{\pi_1,\pi_2}
    = \min_{\pi_2}\max_{\pi_1}V^{\pi_1,\pi_2}.
\end{equation}
The domains studied in this paper are finite, deterministic, perfect-information instances of this setting. Expressing values from the perspective of the player to move reduces their optimality equations to the negamax form in Equation~\ref{eq:bellman-negamax}.

\subsection{AlphaZero and MCTS}

AlphaZero uses a neural network with a value output $V_\theta(s)$ and a policy output $\pi_\theta(a\mid s)$. During self-play, MCTS uses the policy to guide exploration and the value to evaluate leaf nodes. The policy returned by search is derived from root visit counts:
\begin{equation}
    \pi_{\mathrm{MCTS}}(a\mid s)
    = \frac{N(s,a)^{1/\tau}}
    {\sum_b N(s,b)^{1/\tau}},
\end{equation}
where $N(s,a)$ is the visit count of action $a$ and $\tau$ is a temperature parameter. The network is trained from the search policy and the final outcome $z$ using
\begin{equation}
    \mathcal{L}_{\mathrm{AZ}}(\theta)
    = \sum_i \left(V_\theta(s_i)-z_i\right)^2
    - \sum_i \pi_{\mathrm{MCTS}}(\cdot\mid s_i)^\top
      \log \pi_\theta(\cdot\mid s_i).
\end{equation}

Our MCTS uses PUCT~\citep{rosin2011multi}. At an internal node, it selects the action maximizing
\begin{equation}
    Q(s,a)
    + c_{\mathrm{puct}}\,\pi_\theta(a\mid s)
      \frac{\sqrt{N(s)}}{1+N(s,a)},
\end{equation}
where $Q(s,a)$ is the backed-up action value, $N(s)$ is the total visit count of the node, and $c_{\mathrm{puct}}$ controls exploration. During training, Dirichlet noise is added to the root prior to encourage diverse self-play trajectories. Algorithm~\ref{alg:az} summarizes the complete training procedure.

The search-policy target is available immediately after MCTS and is recorded for every visited state, including non-terminal states. The value target only becomes available when the episode terminates. At that point, the final outcome, expressed from the appropriate player's perspective, is assigned to every state-policy pair in the trajectory and the completed triples are added to the replay buffer. Partial trajectories are retained across collection iterations until their episode terminates.

\begin{algorithm}[t]
    \caption{AlphaZero through self-play}
    \label{alg:az}
    \begin{algorithmic}[1]
        \REQUIRE iterations $I$, environments $N_{\mathrm{envs}}$, collection steps $K$, simulations $S$, gradient updates $N_{\mathrm{grads}}$, batch size $B$
        \STATE Initialize policy-value parameters $\theta$, replay buffer $\mathcal{D}$, and $N_{\mathrm{envs}}$ game states and trajectories
        \FOR{$i=1,\ldots,I$}
            \FOR{$k=1,\ldots,K$}
                \FOR{each parallel environment with state $s$}
                    \STATE Run $S$ MCTS simulations guided by $(\pi_\theta,V_\theta)$ to obtain $\pi_{\mathrm{MCTS}}(\cdot\mid s)$
                    \STATE Append $(s,\pi_{\mathrm{MCTS}}(\cdot\mid s))$ to the current trajectory
                    \STATE Select an action from $\pi_{\mathrm{MCTS}}$ and advance the game
                    \IF{the game terminates with outcome $z$}
                        \STATE Assign the corresponding outcome to every pair in the trajectory and add the completed triples to $\mathcal{D}$
                        \STATE Reset the environment and its trajectory
                    \ENDIF
                \ENDFOR
            \ENDFOR
            \FOR{$g=1,\ldots,N_{\mathrm{grads}}$}
                \STATE Sample $B$ triples from $\mathcal{D}$ and update $\theta$ using $\mathcal{L}_{\mathrm{AZ}}$
            \ENDFOR
        \ENDFOR
    \end{algorithmic}
\end{algorithm}

\section{AlphaZero.jl Baseline}
\label{appendix:alphazero_jl}

\begin{figure}[h]
    \centering
    \includegraphics[width=\linewidth]{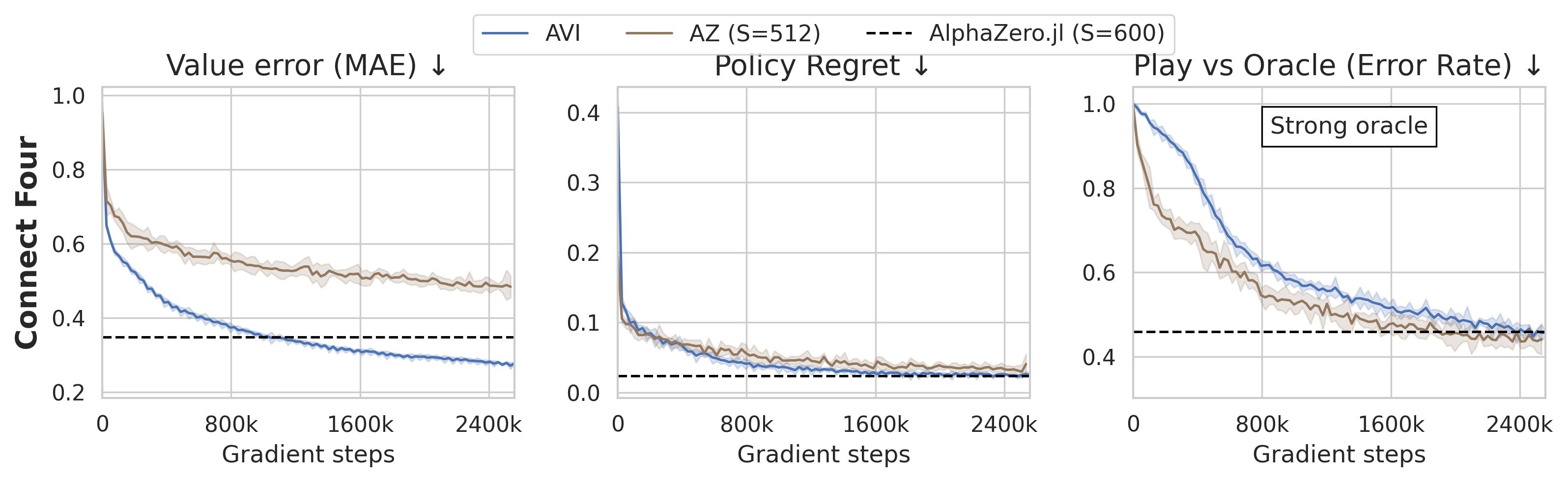}
    \caption{External calibration on Connect Four. We compare AVI and our strongest in-house AlphaZero agent ($S=512$) with AlphaZero.jl ($S=600$). From left to right: value error, policy regret, and error rate against the strong oracle. The AlphaZero.jl result is shown as a horizontal dashed line because only its final checkpoint is evaluated. Differences in architecture, training procedure, and compute budget prevent a controlled comparison; the result is included to calibrate the absolute scale of our in-house AlphaZero performance.}
    \label{fig:alphazero_jl}
\end{figure}

Figure~\ref{fig:alphazero_jl} compares AVI and our strongest in-house AlphaZero configuration with an independently developed Connect Four agent trained using the AlphaZero.jl library~\citep{Laurent2020AlphaZero}. This comparison serves as an external calibration: it indicates that the performance of our in-house AlphaZero agent is not anomalously weak. It is not a controlled algorithmic comparison, because AlphaZero.jl uses a different network architecture, training procedure, and compute budget.

While our primary implementation closely follows the canonical AlphaZero algorithm presented by Silver et al. (2018), the AlphaZero.jl configuration differs in several respects:

\begin{itemize}
    \item \textbf{Averaging positions in the replay buffer:} Instead of treating identically visited states across different games as separate entries, AlphaZero.jl can merge them in the replay buffer. It averages the target policies and values for identical board positions, reducing the noise of the training targets. 
    \item \textbf{Using a convolutional network:} The AlphaZero.jl implementation uses a convolutional neural network (CNN), which has a stronger spatial inductive bias and generally generalizes better in board games. In comparison, our MLP architecture is cheaper to run but less expressive.
    \item \textbf{Decaying the temperature within each game:} Rather than using a fixed exploration temperature $\tau$ or cutting the exploration abruptly after a hardcoded number of moves, the library decreases the temperature parameter smoothly over the course of a single game. This helps the agent naturally transition from exploring early-game openings to deeply exploiting end-game tactics.
    \item \textbf{Replay buffer scheduling:} The maximum capacity of the replay buffer is scheduled to increase dynamically over the course of training, ensuring that the buffer is narrow during early stages to avoid overfitting completely random initial policies, and larger at the end to stabilize learning on high-quality data.
\end{itemize}

These differences may affect both learning efficiency and final performance. We therefore keep AlphaZero.jl separate from the main figure, where AVI and AlphaZero share the same environments, replay buffer, optimization code, and neural-network backbone. The external result is evidence about the scale of our baseline, not evidence for a causal comparison between the learning algorithms.

\section{Experiment Details}
\label{appendix:experimental_details}

In this section, we provide the detailed experimental setup for Connect Four, Hex(7x7), Othello, Go(9x9), and F-Games, including hyperparameters, neural-network architectures, and input representations.

\paragraph{Indicative wall-clock times}
All Connect Four and Hex(7x7) runs used a single GPU, but they were conducted on heterogeneous machines, most commonly with an NVIDIA GTX 1080 Ti. On Connect Four, AVI required approximately two hours of training, while AlphaZero ranged from approximately three hours with $S=32$ to two days with $S=512$. On Hex(7x7), AVI required approximately 20 hours, while AlphaZero ranged from approximately 20 hours with $S=32$ to four days with $S=512$. These measurements provide practical context but are not controlled hardware comparisons; we therefore use the forward-equivalent compute proxy in the main analysis.

\subsection{Connect Four and Hex(7x7)}

\paragraph{Architecture}
For Connect Four (respectively, Hex), we use a neural network consisting of a common body followed by separate value and policy heads. The body contains four (respectively, two) residual blocks, each consisting of a four-layer (respectively, two-layer) Multi-Layer Perceptron (MLP) with residual connections and Layer Normalization. Each layer has 256 hidden units and uses GeLU activations. The value and policy heads are also two-layer MLPs with 256 hidden units. The value head outputs a scalar with no activation function, while the policy head outputs one logit per action.

\paragraph{Input Representation}
The input to the network is a $6 \times 7 \times 2$ binary tensor for Connect Four and a $7\times 7\times 2$ tensor for Hex(7x7). The first channel marks the stones of the current player, and the second marks the stones of the opponent. This tensor is flattened before being fed into the MLP. Both environments have a single value-preserving reflection. To enforce this symmetry, we feed both the original board and its reflected version to the network. The value output is averaged across the two inputs, and the policy output is the average of the original policy and the reflected policy of the symmetric input.

\paragraph{Hyperparameters}
\Cref{tab:vi_params_c4} and \Cref{tab:az_params_c4} list the hyperparameters used for Approximate Value Iteration (AVI) and AlphaZero, respectively, on Connect Four and Hex. For these two games, hyperparameters are shared beyond the necessary changes in network heads. For AVI, we search over $\epsilon\in\{0.05,0.15,0.3\}$ and select $0.3$. For AlphaZero, a parameter search over $c_\text{puct} \in \{1, 2, 3\}$ consistently identifies 3 as the strongest configuration. During evaluation, a lower UCT temperature ($\tau = 0.2$) and a reduced Dirichlet noise fraction ($\epsilon = 0.05$) yield better performance.

\begin{table}[h]
    \centering
    \caption{Hyperparameters for AVI on Connect Four and Hex.}
    \label{tab:vi_params_c4}
    \begin{tabular}{lc}
        \toprule
        Parameter & Value \\
        \midrule
        Number of environments ($N_\text{envs}$) & 128 \\
        Collection steps ($K$) & 128 \\
        Optimizer & Adam \\
        Learning rate & $3 \times 10^{-4}$ \\
        Batch size & 256 \\
        Replay buffer size & 1,000,000 \\
        Epochs per iteration & 4 \\
        Greedy $\epsilon$ & 0.3 \\
        Lookahead depth & 1 \\
        \bottomrule
    \end{tabular}
\end{table}

\begin{table}[h]
    \centering
    \caption{Hyperparameters for AlphaZero on Connect Four and Hex.}
    \label{tab:az_params_c4}
    \begin{tabular}{lc}
        \toprule
        Parameter & Value \\
        \midrule
        Number of environments ($N_\text{envs}$) & 128 \\
        Collection steps ($K$) & 128 \\
        Optimizer & Adam \\
        Learning rate & $3 \times 10^{-4}$ \\
        Batch size & 256 \\
        Replay buffer size & 1,000,000 \\
        Training steps per iteration & 256 \\
        MCTS simulations ($S$) & \{32, 64, 128, 256, 512\} \\
        UCT temperature ($\tau$) (train / eval) & 1.0 / 0.2 \\
        Dirichlet $\alpha$ & 1.0 \\
        Dirichlet fraction $\epsilon$ (train / eval) & 0.25 / 0.05 \\
        PUCT $c_{\text{puct}}$ & 3.0 \\
        \bottomrule
    \end{tabular}
\end{table}

\subsection{Othello and Go(9x9)}
\label{appendix:othello_go}

\paragraph{Game implementations}
We use the Othello and Go(9x9) environments implemented in PGX~\citep{koyamada2023pgx}. Go uses a komi of 7.

\paragraph{Architecture and symmetry augmentation}
MiniZero uses the same board-game network architecture for Othello and Go(9x9): a shared backbone containing three residual blocks, followed by separate policy and value heads~\citep{wu2024minizero}. To make the comparison as direct as possible, our AVI networks use the same backbone and value head, without the policy head that AVI does not require.

Both games admit the eight rotations and reflections of the square board. During AVI training, we randomly sample one of these symmetries for each training example and transform its board representation; the scalar value target remains unchanged. This differs from our treatment of Connect Four and Hex(7x7), where symmetry is enforced by evaluating symmetric versions of each state and averaging their outputs. Thus, the Othello and Go(9x9) experiments use standard data augmentation rather than an explicitly equivariant inference procedure.

\paragraph{Training}
We leave the AVI data-collection and optimization procedure unchanged from Connect Four and Hex(7x7), using the settings in Table~\ref{tab:vi_params_c4}. We only reduce the exploration rate from $\epsilon=0.3$ to $\epsilon=0.15$ to account for the larger action spaces. We train five independent seeds for $5.12$M gradient steps on NVIDIA H100 GPUs.

\paragraph{MiniZero baseline and evaluation}
We use the publicly released MiniZero AlphaZero agents for Othello and Go(9x9), whose networks were trained with 200 MCTS simulations per move~\citep{wu2024minizero}. We also use 200 simulations during evaluation. Native MiniZero inference uses its policy and value heads, while cross-inference keeps MiniZero's policy prior and replaces its leaf value by AVI's prediction. As explained in Section~\ref{sec:match_evaluation}, this diagnostic configuration evaluates two separate backbones and is not intended as an efficient agent.

We evaluate AVI at $0$, $1.28$M, $2.56$M, $3.84$M, and $5.12$M gradient steps. For Othello, we use all 54 positions attainable after three plies; for Go(9x9), we use all 81 positions attainable after the first move. Every opening is played from both sides, resulting in 108 Othello games and 162 Go(9x9) games per checkpoint, inference method, and AVI seed. We average the resulting scores over the five seeds and report 95\% confidence intervals using Student's $t$ distribution.

\subsection{F-Games}

\paragraph{Architecture}
For F-Games, we use the residual MLP design of Connect Four and Hex(7x7), with a body containing two residual blocks. Each block uses Layer Normalization and 256 hidden units. The body is followed by separate value and policy heads.

\paragraph{Input Representation}
The input for F-Games is the sequence of actions played so far. Each action is one-hot encoded, and future (unknown) actions are encoded using a special ``unknown'' token, increasing the vocabulary size by 1. The input sequence length is fixed to $h \times (b+1)$, covering the maximum possible game length. This sequence is flattened and fed into the two-block residual MLP described above.

\paragraph{Hyperparameters}
\Cref{tab:vi_params_fgames} and \Cref{tab:az_params_fgames_app} list the hyperparameters used for AVI and AlphaZero on F-Games. For AVI, we reduce the number of epochs per iteration because we found that too many gradient updates were detrimental to stability in this domain. For AlphaZero, we retain the other hyperparameters but reduce the collection and gradient steps to reflect the shorter horizon.

\begin{table}[h]
    \centering
    \caption{Hyperparameters for AVI on F-Games.}
    \label{tab:vi_params_fgames}
    \begin{tabular}{lc}
        \toprule
        Parameter & Value \\
        \midrule
        Number of environments ($N_\text{envs}$) & 4096 \\
        Collection steps ($K$) & 128 \\
        Optimizer & Adam \\
        Learning rate & $3 \times 10^{-4}$ \\
        Batch size & 256 \\
        Replay buffer size & 1,000,000 \\
        Epochs per iteration & 1 \\
        Greedy $\epsilon$ & 0.3 \\
        \bottomrule
    \end{tabular}
\end{table}

\begin{table}[h]
    \centering
    \caption{Hyperparameters for AlphaZero on F-Games.}
    \label{tab:az_params_fgames_app}
    \begin{tabular}{lc}
        \toprule
        Parameter & Value \\
        \midrule
        Number of environments ($N_\text{envs}$) & 4096 \\
        Collection steps ($K$) & 32 \\
        Optimizer & Adam \\
        Learning rate & $3 \times 10^{-4}$ \\
        Batch size & 1024 \\
        Replay buffer size & 1,000,000 \\
        Training steps per iteration & 200 \\
        MCTS simulations ($S$) & \{32, 64, 128, 256\} \\
        UCT temperature ($\tau$) & 1.0 \\
        Dirichlet $\alpha$ & 1.0 \\
        Dirichlet fraction $\epsilon$ & 0.25 \\
        PUCT $c_{\text{init}}$ & 1.25 \\
        PUCT $c_{\text{base}}$ & 19652 \\
        \bottomrule
    \end{tabular}
\end{table}

\section{Connect Four}
\label{appendix:connect_four}

\subsection{Game Rules}
Connect Four is a classic two-player connection board game played on a vertically suspended grid consisting of 7 columns and 6 rows. Players designate a color and then alternately drop one of their colored discs into any of the non-full columns. The discs fall straight down, occupying the lowest available space within the column. The objective of the game is to be the first to form a contiguous line of four of one's own discs, either horizontally, vertically, or diagonally. If the board fills up completely before either player achieves this goal, the game ends in a draw. The state space of Connect Four contains approximately $4.53 \times 10^{12}$ possible positions.

\subsection{Solver details}
To evaluate our agents against a ground truth, we rely on a popular and highly optimized Connect Four solver developed by Pascal Pons.\footnote{\url{https://github.com/PascalPons/connect4}} This solver is based on an alpha-beta search algorithm, extended with a transposition table and an opening book, making it extremely efficient for exact resolution. 

Crucially, in contrast to the Hex oracle, this Connect Four solver constitutes a ``strong'' oracle. It does not merely output whether a position is a theoretical win, loss, or draw; it also provides the exact distance to the outcome (the minimum number of moves required to force a win, or the maximum number of moves to delay a loss). It can therefore distinguish actions with the same theoretical outcome, making the play-vs-oracle evaluations particularly strict.

\subsection{Datasets}

To compute balanced distributions of states for the regret- and value-error-based evaluations, we systematically sample states in two distinct ways.

(1) Uniform dataset: for this dataset, we collected 1500 states from uniform random play. This yields wildly diverse positions, but often concentrated in the early game, since random play tends to end quickly. To ensure data diversity, we classify the states by the number of moves played and the number of remaining moves, and we stratify the sampling to get a balanced distribution across these two dimensions. This dataset is used for the main evaluation in Figure~\ref{fig:az_vs_vi}. (2) $\epsilon$-optimal dataset: for this dataset, we collected 1500 states by making the solver play against itself, but with an $\epsilon$-greedy strategy. This dataset is more concentrated on states that are likely to be encountered during actual play, as it simulates a more realistic distribution of positions.

Each dataset is further divided into six subsets based on the number of moves played and remaining, following the methodology of \citet{Laurent2020AlphaZero,ponssolver}. These two dimensions are useful proxies for the game phase and difficulty of a position. The resulting subsets are as follows:

\begin{itemize}
    \item \textbf{Easy Opening Positions:} These positions have 14 or fewer moves played, with fewer than 14 moves remaining. They represent straightforward scenarios where the optimal strategy is relatively simple to discern.
    \item  \textbf{Medium Opening Positions:} These positions also have 14 or fewer moves played but have between 14 and 28 moves remaining. They present a moderate level of complexity, requiring more strategic foresight.
    \item \textbf{Hard Opening Positions:} These positions have 14 or fewer moves played, with 28 or more moves remaining. They are the most challenging opening scenarios, demanding deep strategic planning to navigate effectively.
    \item \textbf{Easy Midgame Positions:} These positions have between 14 and 28 moves played, with fewer than 14 moves remaining. They represent midgame scenarios where the game is still relatively open.
    \item \textbf{Medium Midgame Positions:} These positions have between 14 and 28 moves played, with between 14 and 28 moves remaining.
    \item \textbf{Easy Endgame Positions:} These positions have 28 or more moves played, with fewer than 14 moves remaining. They represent endgame scenarios where the outcome is often more predictable.
\end{itemize}

\subsection{Detailed results}
\label{appendix:additional_results_c4}

In this section, we provide a detailed breakdown of the performance of Approximate Value Iteration and AlphaZero on Connect Four.

\begin{figure}[h]
    \centering
    \includegraphics[width=0.8\linewidth]{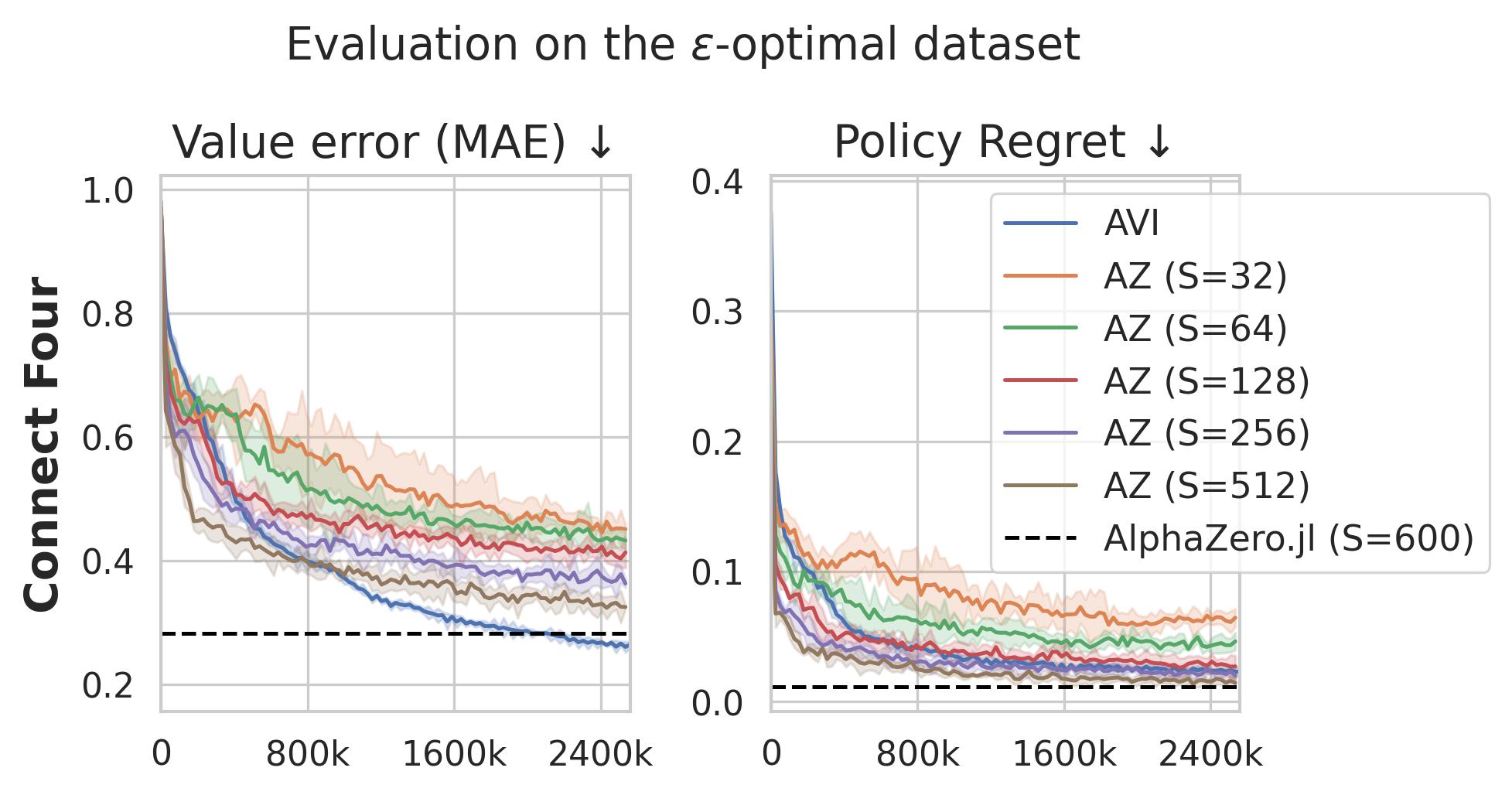}
    \caption{Value error (MAE) and policy regret on the $\epsilon$-optimal dataset for Connect Four. AVI curves average 20 random seeds and AlphaZero curves five seeds. Shaded regions show 95\% confidence intervals computed with Student's $t$ distribution.
    }
    \label{fig:greedy_dataset}
\end{figure}

\paragraph{$\epsilon$-optimal dataset} Figure~\ref{fig:az_vs_vi} reports value error and policy regret on the uniform dataset. Figure~\ref{fig:greedy_dataset} provides the corresponding evaluation on the $\epsilon$-optimal dataset.

\begin{figure}[h]
    \centering
    \includegraphics[width=\linewidth]{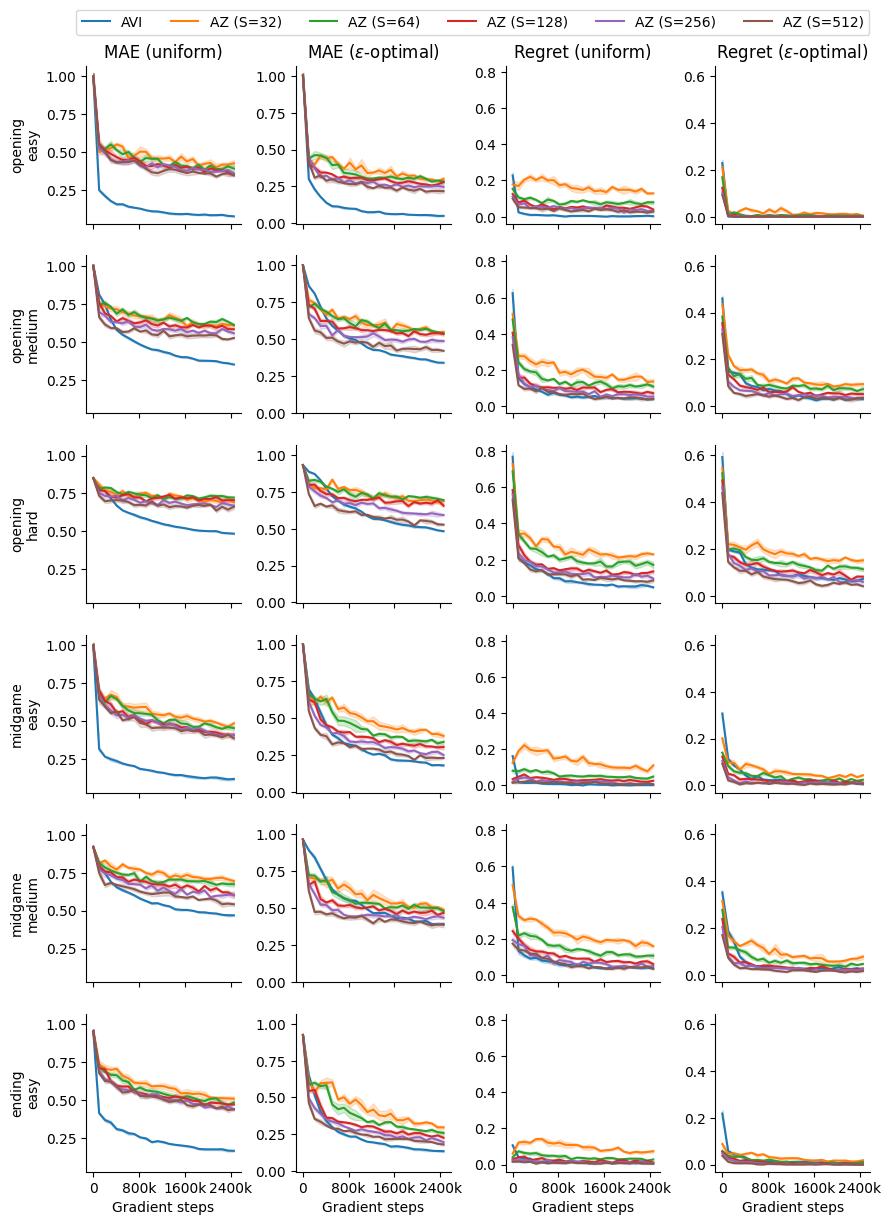}
    \caption{Breakdown of the value error (MAE) and policy regret on the different Connect Four datasets. AVI consistently achieves lower or competitive error rates across all game stages and difficulty levels compared to AlphaZero variants, confirming that the bootstrapping approach yields finer value estimates even in complex positions. AVI curves average 20 random seeds and AlphaZero curves five seeds. Shaded regions show 95\% confidence intervals computed with Student's $t$ distribution.
    }
    \label{fig:az_vs_vi_value_breakdown}
\end{figure}

\paragraph{Breakdown by state type} As described in the last section, our evaluation relies on a suite of datasets covering different game phases (Opening, Midgame, Endgame) and difficulty levels (Easy, Medium, Hard). Figure~\ref{fig:az_vs_vi_value_breakdown} displays the evolution of the value error and policy regret for each of these specific subsets. AVI curves average 20 random seeds and AlphaZero curves five seeds; shaded regions show 95\% confidence intervals computed with Student's $t$ distribution.
\clearpage

\section{Hex}
\label{appendix:hex}

\subsection{Game Rules}
Hex is a two-player perfect information board game played on a hexagonal grid. In our experimental setting, we consider the 7x7 version of the game. Two players take turns placing stones of their respective colors on empty cells of the grid. The objective for each player is to form an unbroken chain of their stones connecting their two opposing sides of the board. Due to the topology of the hexagonal grid, the game cannot end in a draw; one player will inevitably complete a winning chain. 

\subsection{Solver details}
While Hex ($7\times 7$) is mathematically solved, its state space of roughly $10^{22}$ positions makes it intractable for standard unoptimized minimax search. To provide ground-truth evaluations for our AI agents, we rely on the MoHex engine, which features a highly optimized perfect solver for $7\times 7$ boards. MoHex utilizes complex domain-specific reasoning, such as computing virtual connections, inferior cell analysis, and topology properties, combined with depth-first proof-number search. This allows it to find the exact value of most positions in less than a minute on average.

As noted in the main text, one limitation of this solver compared to our Connect Four solver is that MoHex only outputs the theoretical game-theoretic value of a position (Win or Loss). It does not maintain a notion of distance to the outcome (i.e., which move wins the quickest or loses the slowest). Consequently, in our play-vs-oracle evaluations, the Hex oracle acts as a "weak" oracle: it evaluates all winning (or losing) moves as strictly equal.

\subsection{Datasets}
Because the MoHex solver is computationally expensive, it does not allow us to evaluate and categorize states with the same high granularity as in Connect Four. Therefore, for our empirical evaluations, we rely on a single evaluation dataset composed of 445 distinct positions sampled via uniform random play.

For the $\mathcal{D}_\text{openings}$ dataset used in our play-vs-oracle evaluation, we use 128 positions after exactly two plies, balanced between 64 winning and 64 losing positions according to the solver.

\section{F-Games}
\label{appendix:f_games_generation}

In this section, we describe the generation process for the synthetic F-Games used in our experiments. These games are based on the \textit{forward model} proposed by \citep{boige2025alphabeta}, which allows for the procedural generation of game trees with known minimax values.

\subsection{Game generation protocol}
Unlike traditional random game-tree models that sample leaf values independently and propagate them upwards (minimax), F-games are constructed top-down. The key insight is to sample children values conditionally on the parent's value to ensure the minimax consistency is preserved at every step. This method allows generating games with a controlled distribution of values while providing the ground-truth value of every node in the tree explicitly.

Formally, let $\mu$ be a discrete distribution over a value support $\mathcal{V} = \{-n, \dots, n\}$. The generation process begins by sampling the root value $v_{root} \sim \mu$. Then, for any node with an assigned value $v$ at height $h > 0$, its children values $v_1, \dots, v_b$ are generated as follows:

\begin{enumerate}
    \item A ``special child'' index $k$ is selected uniformly at random from $\{1, \dots, b\}$.
    \item This child is assigned the negation of the parent's value: $v_k = -v$. This ensures that the negamax constraint $\max_i (-v_i) \ge v$ holds.
    \item The remaining $b-1$ children are sampled independently from the distribution $\mu$, restricted to the interval $[-v, n]$. Specifically, if $X \sim \mu$, then for $j \neq k$, $v_j \sim X | X \ge -v$. This ensures that $-v_j \le v$, satisfying the condition $\max_i (-v_i) = v$.
\end{enumerate}

This procedure is applied recursively until the leaf nodes are reached.
\subsection{Unbalanced Trees}
To increase the realism of the generated games, we introduce a probability of premature termination $\beta$. During the recursive generation, at any node that is not at the maximum depth $h$, the generation process halts with probability $\beta$, turning the node into a terminal leaf with its assigned value. This results in unbalanced trees where branches can have varying lengths, mimicking the structure of real games where game-ending states can occur before the maximum game length.

\subsection{Experiment setting}

 In all experiments, $n$ is set to 1, meaning we consider values in $\{-1, 0, 1\}$, corresponding to loss, draw, and win outcomes. The value distribution $\mu$ is selected as the uniform distribution over $\mathcal{V}$, a natural choice that does not bias the game towards any particular outcome. We always set the root value to $1$; this ensures the first player has a winning strategy and the game is not an all-draw game. The probability of unbalanced tree termination is uniformly set to $\beta=0.05$.

To investigate the scalability and consistency of Value Iteration across varying algorithmic complexities, we configure our F-games across three distinct tree structures:
\begin{itemize}
    \item \textbf{Deep and narrow:} Height $h=20$ and branching factor $b=2$.
    \item \textbf{Balanced:} Height $h=10$ and branching factor $b=5$.
    \item \textbf{Shallow and wide:} Height $h=5$ and branching factor $b=20$.
\end{itemize}

\begin{figure}[ht]
    \centering
    \includegraphics[width=\linewidth]{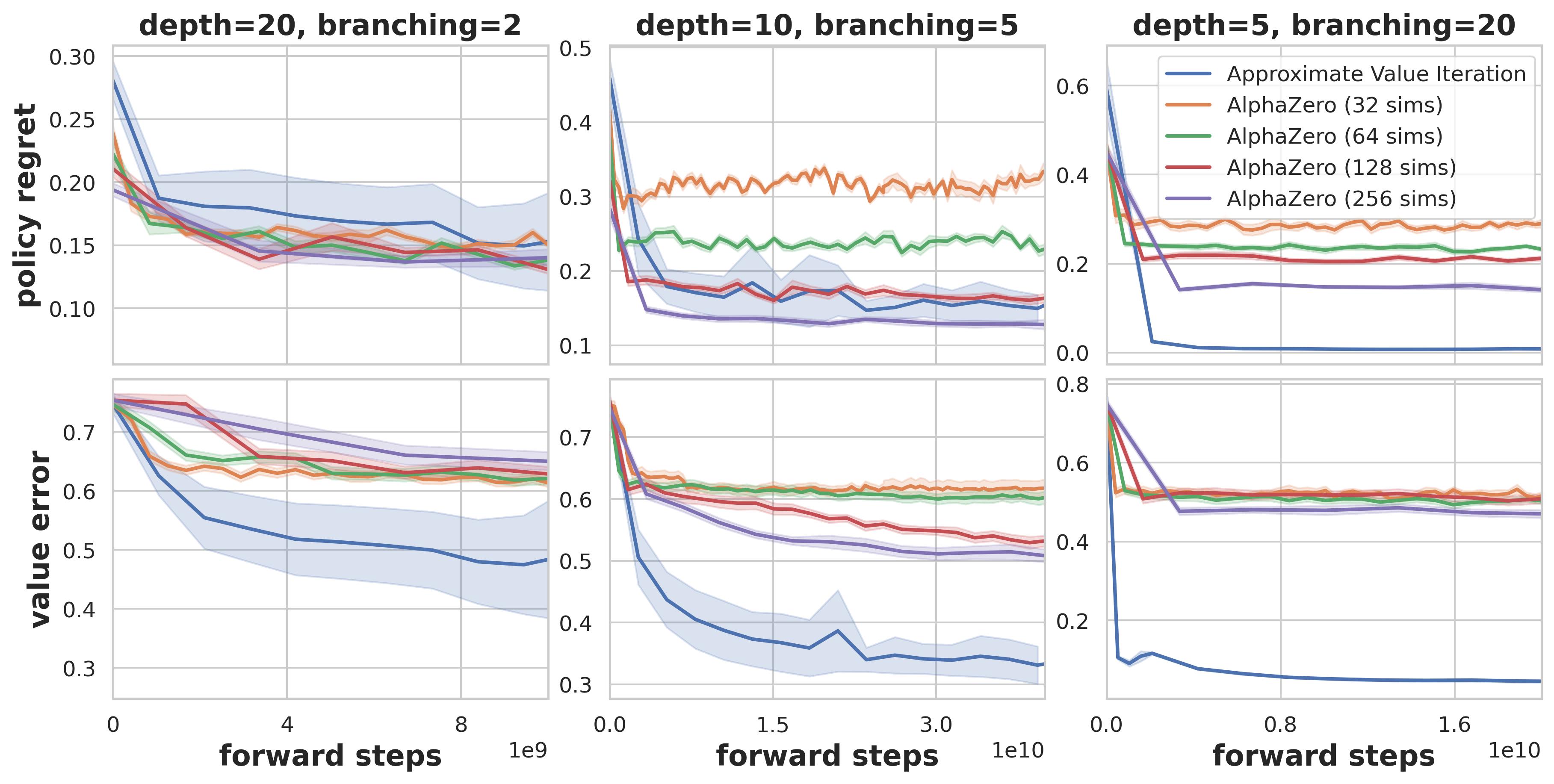}
    \caption{Comparison of Value Iteration and AlphaZero on F-Games. We report the Mean Absolute Error of the value function (top row) and the Policy Regret (bottom row) for three different game structures: deep and narrow ($h=20, b=2$), balanced ($h=10, b=5$), and shallow and wide ($h=5, b=20$). The x-axis represents the computational cost measured in neural network forward passes. Shaded areas indicate the standard error over 10 random game instances.}
    \label{fig:fgame_az_vs_vi}
\end{figure}

For each of these configurations, we independently generate 10 game instances to average out the variance related to a particular seed's topology and outcome distribution. The corresponding results are depicted in Figure~\ref{fig:fgame_az_vs_vi}.

\subsection{Datasets}
To compute the policy regret and value error for these synthetic domains, we extract an evaluation set for each game instance. Specifically, we unroll a fully random uniform policy within the game environment to visit diverse trajectories and sample a robust dataset of 4096 distinct valid states per F-Game.

\end{document}